\documentclass[a4paper]{article}

\usepackage[english]{babel}
\usepackage[utf8x]{inputenc}
\usepackage[T1]{fontenc}

\usepackage{algorithm}
\usepackage{algorithmic}
\usepackage{enumitem} 
\usepackage{amsmath}
\usepackage{amssymb}
\usepackage{graphicx}
\usepackage[colorinlistoftodos]{todonotes}
\usepackage[colorlinks=true, allcolors=blue]{hyperref}
\usepackage{color}
\usepackage{multirow}
\usepackage[round, sort,comma,authoryear]{natbib}
\usepackage{siunitx} % for exponential notation
\usepackage{verbatim}
\usepackage{tikz}

\newcommand{\footnoteremember}[2]{
\footnote{#2}
  \newcounter{#1}
  \setcounter{#1}{\value{footnote}}
}
\newcommand{\footnoterecall}[1]{
\footnotemark[\value{#1}]
}

\title{Predicting Tactical Solutions to Operational Planning Problems under Imperfect Information}

\author{
Eric Larsen\footnoteremember{cirrelt}{Department of Computer Science and Operations Research and CIRRELT, Universit\'e de Montr\'eal}
\and
S\'ebastien Lachapelle\footnoteremember{Lachapelle}{Department of Computer Science and Operations Research and Mila, Universit\'e de Montr\'eal}
\and
Yoshua Bengio\footnoteremember{Bengio}{Department of Computer Science and Operations Research and Mila, Universit\'e de Montr\'eal, CIFAR Senior Fellow}
\and
Emma Frejinger\footnoterecall{cirrelt}\footnote{Corresponding author. Email: frejinger.umontreal@gmail.com}
\and
Simon Lacoste-Julien\footnoteremember{Lacoste}{Department of Computer Science and Operations Research and Mila, Universit\'e de Montr\'eal, CIFAR Fellow}
\and
Andrea Lodi\footnoteremember{Lodi}{Canada Excellence Research Chair, Polytechnique Montr\'eal}}

\date{\today}

\begin{document}

\maketitle

\begin{abstract}
This paper offers a methodological contribution at the intersection of machine learning and operations research. Namely, we propose a methodology to quickly predict expected tactical descriptions of operational solutions (TDOSs). The problem we address occurs in the context of two-stage stochastic programming where the second stage is demanding computationally. We aim to predict at a high speed the expected TDOS associated with the second stage problem, conditionally on the first stage variables. This may be used in support of the solution to the overall two-stage problem by avoiding the online generation of multiple second stage scenarios and solutions. We formulate the tactical prediction problem as a stochastic optimal prediction program, whose solution we approximate with supervised machine learning. 
The training dataset consists of a large number of deterministic operational problems generated by controlled probabilistic sampling. The labels are computed based on solutions to these problems (solved independently and offline), employing appropriate aggregation and subselection methods to address uncertainty. Results on our motivating application on load planning for rail transportation show that deep learning models produce accurate predictions in very short computing time (milliseconds or less).
The predictive accuracy is close to the lower bounds calculated based on sample average approximation of the stochastic prediction programs.
\end{abstract}


\section{Introduction} \label{sec:intro}

Operations research (OR) has been successful in developing methodologies and algorithms to solve efficiently various types of decision problems that can be formalized but are nevertheless too complex or time-consuming for humans to process. These methodologies and algorithms are crucial to a wealth of applications. Conversely, machine learning (ML) and its subfield known as deep learning (Goodfellow et al. 2016) have had remarkable success in automating tasks that are easy to accomplish but difficult to formalize by humans, for example, image analysis, natural language processing, voice and face recognition. Through this undertaking, ML has developed an array of powerful classification and regression methods that can be used to approximate generic input-output maps and perform conditional predictions under uncertainty. Building on those strengths, we propose a methodology where OR and ML complement each other and are respectively applied to the tasks they
are best suited for in order to address a conditional prediction problem that could not otherwise be solved within the allowed time budget.

The problem we address occurs in the context of two-stage stochastic programming where the second stage is demanding computationally. We focus on the second stage. Namely, we aim to predict at a high speed the expected solution of the second stage problem, conditionally on the first stage variables. This may be used in support of the solution to the overall two-stage problem by avoiding online scenario generation. Our predictions stem from supervised ML. They account for the stochasticity in the second stage problem and can approximate its expected solution at a higher online speed than that which can be achieved with alternative methods.

As a motivating application, we consider a myopic, two-stage booking control problem pertaining to cargo capacity management. The first stage \emph{tactical} problem consists in maximizing expected profit by controlling the accept/reject decision of a booking request for the transportation of a specified number of objects of standardized types on board available vehicles of standardized types. At the time of booking, uncertainty may surround the exact characteristics of the future shipment. This uncertainty is to be resolved in the second stage, at the time of loading. At this point, given complete knowledge of the exact characteristics of the shipment and available vehicles, an NP-hard \emph{operational} multidimensional bin packing problem must be solved.

Calculating the net profit resulting from a particular loading solution does not require the fully detailed assignment of individual objects to individual positions in vehicles returned by the operational solution. It is sufficient to know the number of objects of each type that are loaded and the number of vehicles of each type that are used. We call such a characterization of the operational solution useful from the tactical standpoint a \emph{tactical description of operational solution} (TDOS). The level of detail of a TDOS is lesser than that featured in the operational solution since its role is to support early planning decisions rather than to set out a detailed load plan. In comparison, a strategic description of operational solution would comprise even fewer details, such as the value of the objective function achieved by the solution of the operational problem. The idea of using a less detailed solution representation at the tactical level is not new. Instead, the novelty here lies in the proposed methodology that combines ML and discrete optimization. We note that our methodology could also be applied in the context of the general sequential booking problem \citep{LeviEtAl11, BarzEtAl16} to the prediction of the expected value achieved in the operational loading problem.

The idea underlying the proposed methodology is simple and attractive: predict the expected TDOS, conditionally on first stage variables, using supervised ML, where the training data consists of a large number of deterministic operational problems that have been solved independently and offline. First, operational problem instances under perfect information (the input) are sampled from the space relevant to the application at hand and solved using an existing deterministic optimization model and a discrete optimization solver. Second, the detailed operational solutions are synthesized according to the chosen TDOS (the output). Third, an ML approximator is trained based on the generated input-output data while accounting for imperfect information regarding certain features of the input. Fourth, this approximator is used to generate predictions of the expected TDOS based on information available at tactical standpoint. These predictions are expected to be delivered with high speed, high accuracy, as well as low marginal cost in terms of data, memory and computing requirements.

Some important challenges arise in this undertaking. First, the structure of the output is tied to the chosen TDOS and can be of fixed or variable size. Second, inputs and outputs defined in the chosen solution description may be related by a number of explicit constraints that must also be satisfied by the predictions. Third, missing information regarding a subset of the inputs needs to be addressed through appropriate sampling, subselection and aggregation methods. The designs of the ML models and algorithms depend on how these challenges are met.

To illustrate the methodology, we specialize the two-stage booking problem to reflect the booking decisions made about the shipment of intermodal containers by rail at our industrial partner. In this case, the operational bin packing problem is the so-called load planning problem (LPP) where containers are loaded optimally onto double-stack railcars \citep{MantEtAl17}. The latter cannot be solved deterministically at the time of booking due to imperfect information about container weights. The TDOS consists in capacity usage, that is, in the numbers of containers of each type that would be loaded and the numbers of railcars of each type that would be used, if the booking request were accepted.

Given installed capacity (i.e., the numbers of railcars of each type known in advance to be available at loading time), a risk-neutral manager accepts a booking demand if the associated expected profit (transportation fees received upfront minus expectation of the future cost of loading) is positive. The future cost of loading is a linear function of the number of containers whose booking had been accepted but that must be left behind (in case of insufficient capacity) and of the number of railcars actually used for loading. This cost is realized at operational loading time, after the uncertainty in container weights has been resolved, and can be computed for each particular realization of the weights from the statement of the booking request and the TDOS. The expected future cost of loading can be computed from the statement of the booking request and the expected TDOS. In addition, the advance information embodied in the expected TDOS can be used to plan the railcar and container layouts in the terminal.

We examine specifically the problem of predicting at high speed - in real time - with a trained ML approximator the expected TDOS pertaining to the railway container booking problem, conditionally on a booking request and a set of available railcars. This problem is of practical importance and features characteristics that make it useful for illustrating the proposed methodology: On the one hand, an integer linear programming (ILP) formulation of the operational LPP problem can be solved under full information by commercial ILP solvers within seconds or minutes (Mantovani et al. 2018). On the other hand, this formulation cannot be used for the application due to advance information about container weights being usually unavailable, and to computing time being too long to support a real-time application.

\paragraph{Short Literature Review and Pointers.}

The application of ML to discrete optimization problems was the focus of an important research effort in the 1980's and 1990's \citep{Smit99}. However, limited success was ultimately achieved and this area of research was left nearly inactive from the beginning of this century. A renewal of interest has been kindled by the successes witnessed in deep learning and the state of the art is advancing at an increasing pace. 

The most successful locus of synergies between OR and ML, which has attracted a huge amount of attention in the last ten years, is the introduction of continuous optimization methods originating in OR to algorithms used in ML, most notably the stochastic gradient descent algorithm. The interested reader is referred to, e.g., \cite{CS17} for a tutorial.

More recent is the (mainly) exploratory bulk of work devoted to the use of ML for discrete optimization problems. On the one hand, ML is used as a tool for approximating complex and time-consuming tasks in OR algorithms, arguably the most important one being branching for enumerative approaches to NP-hard problems \citep[see, e.g.,][for a survey]{LodiZarp17}.
On the other hand, ML is used to more directly solve (of course, heuristically) discrete optimization problems. A number of ML methodologies can be used for this purpose. Although we are not proposing a heuristic solution to the LPP, some methodologies in this area are closer to our own. Those are based on supervised learning, such as, for example, \cite{VinyEtAl15} that focuses on predicting fully detailed solutions to the famous Traveling Salesman Problem. However, those supervised learning methodologies address \emph{deterministic} problems. The interested reader is referred to \cite{BLP18} for a very recent survey on all aspects of ML for discrete optimization. Finally, there is a growing interest for the application of ML to the solution of stochastic optimization problems. In this area of research, we note for instance \cite{NairEtAl18, BengEtAl19, AbbaEtAl20}.

\paragraph{Paper Contributions.}

The paper offers four main contributions.
\begin{enumerate}
    \item
    We address a problem occurring in the context of two-stage stochastic programming where the second stage is demanding computationally. We concentrate on the high-speed prediction of the \emph{expected} solution to the second stage problem, conditionally on the first stage variables. 
Hence, we formulate a stochastic optimal prediction problem (Section~\ref{sec:pbDescription}), whose solution we approximate with supervised ML (Section~\ref{sec:methodology}). The approximations generated by ML are intended to support the solution of the overall two-stage problem by avoiding the online generation of multiple second stage scenarios and solutions.
        
     Our proposal is robust to the caveat expressed in \cite{Wall00}: We do not seek for a solution to the overall two-stage problem in the loosely defined space spanned by the individual deterministic solutions that can be computed by collapsing the first and second stage for every realization of the stochastic elements in the second stage. Instead, we show how to approximate with ML the expected TDOS associated with the second stage problem, conditionally on the first stage variables. Our purpose is to circumvent the generation of multiple second stage scenarios and solutions that is customary in stochastic programming. 
     
    \item With respect to the current literature using ML for OR, the proposed methodology combines discrete optimization and ML in an innovative way by integrating an ML predictor / approximator to deal with the data uncertainty that is inherent to the strategic and tactical planning levels.
    
    Crucially, by omitting from the ML input data the variables that are unknown at prediction standpoint, whereas these variables are sampled in the dataset and used in calculating the output data used for ML, the trained approximator resulting from ML optimizes the empirical statistical risk associated with these unknown variables. By selecting the mean squared error as the underlying loss function, the ML approximator returns approximate expected values.
    
    Deterministic settings may be viewed as special cases. For example, \cite{FiscFrac17} consider, at the strategic level, a deterministic wind farm layout optimization problem and use ML to predict the objective values achieved by candidate sites.
    
    \item In contrast with existing alternatives offered by approximate stochastic programming such as sample average approximation (SAA), response surface, stochastic approximation, stochastic search, our methodology built on ML anticipates calculations by generating, in advance, a \emph{prediction function} instead of pointwise solutions on demand. Hence, it does not require to solve deterministic operational problems at prediction time.
    
    We illustrate through an extensive computational evaluation on our real-world application how this comparative advantage allows to build both fast and accurate predictors. Computing time is in the order of milliseconds whereas accuracy approaches the lower bound calculated based on SAA.
    
    \item Our methodology relies on existing ML models and algorithms. This is a key advantage since we can benefit from the recent advances in this field, in our case deep learning. The methodology leads to state-of-the-art results, i.e., solves a problem that otherwise would have been unsolvable to that extent of accuracy in the allowed time budget, essentially online.
\end{enumerate} 

The remainder of the paper is structured as follows: Section~\ref{sec:pbDescription} defines the prediction problem under consideration and discusses existing solution methods from the field of stochastic programming. Section~\ref{sec:methodology} delineates the proposed methodology whereas Section~\ref {sec:application} presents a detailed application and reports the results. Finally, Section~\ref{sec:conclusion} summarizes the content of the paper, reviews outstanding issues and describes directions for future research.

\section{The Prediction Problem} \label{sec:pbDescription}
The formal prediction problem we are addressing is as follows. Let a particular instance of an operational (deterministic) optimization problem be represented by the input feature vector ${\textbf{x}}$. Optimal operational solutions (i.e., those containing values of all decision variables) are ${\mathbf{y}}^*({\mathbf{x}})
 :\equiv \arg \inf_{{\mathbf{y}} \in \mathcal{Y}({\mathbf{x}})} C({\mathbf{x}}, {{\mathbf{y}}})$, where $C({\mathbf{x}}, {\mathbf{y}})$ and $\mathcal{Y}({\mathbf{x}})$ denote the cost function and the admissible space, respectively. Ahead of the time at which the operational problem is solved, we wish to predict certain characteristics of the optimal operational solutions, based on currently available information. We call such a characterization a TDOS. Information on a subset of the feature vector ${\textbf{x}}$ may be unavailable or incomplete at the time of prediction and we define the partition ${\textbf{x}} = [{\textbf{x}}_{\text{a}}, {\textbf{x}}_{\text{u}}]$ accordingly, where ${\textbf{x}}_{\text{a}}$ contains available features and ${\textbf{x}}_{\text{u}}$ unavailable or yet unobserved ones. This partition is the same for all instances. Furthermore, we denote by $g(.)$ the mapping from the fully detailed operational solution to the TDOS featuring the level of detail relevant to the context at hand. Hence, $g({\mathbf{y}})$ is the synthesis of the operational solution ${\mathbf{y}}$ according to the TDOS embedded in $g(.)$.

 Our goal is to compute or at least approximate the solution $\bar{\mathbf{y}}^*({\mathbf{x}}_{\text{a}})$ to the following optimal prediction stochastic programming  \citep[see, e.g.,][]{KallWall94, BirgLouv11, ShapEtAl09} problem:
 
 \begin{equation}
\bar{\mathbf{y}}^*({\mathbf{x}}_{\text{a}}) : \equiv \arg \inf_ {\bar{\mathbf{y}} \in \bar{\mathcal{Y}}({\mathbf{x}}_{\text{a}})} \Phi_{{\mathbf{x}}_{\text{u}}} \lbrace \lVert
\bar{\mathbf{y}} - g({\mathbf{y}}^*({\mathbf{x}}_{\text{a}}, {\mathbf{x}}_{\text{u}})) \rVert \mid {\mathbf{x}}_{\text{a}} \rbrace
\label{prediction first stage}
\end{equation}

\begin{equation}
{\mathbf{y}}^*({\mathbf{x}}_{\text{a}}, {\mathbf{x}}_{\text{u}}) :\equiv \arg \inf_{{\mathbf{y}} \in \mathcal{Y}({\mathbf{x}}_{\text{a}}, {\mathbf{x}}_{\text{u}})} C({\mathbf{x}}_{\text{a}}, {\mathbf{x}}_{\text{u}}, {\mathbf{y}})
\label{prediction second stage}
\end{equation}
where $\lVert \rVert$ denotes a suitable norm (e.g., the $L_1$- or $L_2$-norm when the output has fixed size) and $\Phi_{{\mathbf{x}}_{\text{u}}} \lbrace \lVert .\rVert \mid {\mathbf{x}}_{\text{a}} \rbrace$ denotes either the expectation or a quantile (e.g., the median) operation over the distribution of ${\mathbf{x}}_{\text{u}}$, conditional upon ${\mathbf{x}}_{\text{a}}$. So, $\bar{\mathbf{y}}^*({\mathbf{x}}_{\text{a}})$ are the optimal predictions of the synthesis of the operational optimizer $g({\mathbf{y}}^*({\mathbf{x}}_{\text{a}}, {\mathbf{x}}_{\text{u}}))$, conditionally on information available from tactical standpoint. Finally, $\mathcal{Y}({\mathbf{x}}_{\text{a}}, {\mathbf{x}}_{\text{u}})$ is the admissible space defined by the set of constraints relevant to the operational context, whereas $\bar{\mathcal{Y}}({\mathbf{x}}_{\text{a}})$ is defined only by the set of constraints relevant to the tactical context. At first sight, the problem might appear as a classic two-stage stochastic program. In contrast to such an optimal control problem, the two stages in our optimal prediction problem are decoupled: the solution to the deterministic problem in the second stage (operational solution) cannot be seen as a recourse to the prediction at the first stage (tactical solution).

In real-time or repeated applications, we need to generate solutions to \eqref{prediction first stage} and \eqref{prediction second stage} at a high speed for any value of ${\mathbf{x}}_{\text{a}}$. Whenever closed-form solutions are unavailable, which usually occurs, it is generally prohibitive to compute a solution to \eqref{prediction first stage} and \eqref{prediction second stage} on demand for every particular value of ${\mathbf{x}}_{\text{a}}$ encountered. As detailed in Section 3, our methodology generates a prediction function that can take any value of ${\mathbf{x}}_{\text{a}}$ as input and outputs accurate predictions $\widehat{\mathbf{y}}^*({\mathbf{x}}_{\text{a}})$ of $\bar{\mathbf{y}}^*({\mathbf{x}}_{\text{a}})$. The predictions are given by $\widehat{\mathbf{y}}^*({\mathbf{x}}_{\text{a}}) \equiv f({\textbf{x}}_{\text{a}};\boldsymbol{\theta})$ where $f(\cdot;\cdot)$ is a particular ML model and $\boldsymbol{\theta}$ is a vector of parameters.

\paragraph{Solutions from Stochastic Programming.} A number of alternative approaches and methods are available from the field of stochastic programming to address the problem defined by \eqref{prediction first stage} and \eqref{prediction second stage}. For general specifications of $C({\mathbf{x}}_{\text{a}}, {\mathbf{x}}_{\text{u}}, {\mathbf{y}})$ and $\mathcal{Y}({\mathbf{x}}_{\text{a}}, {\mathbf{x}}_{\text{u}})$, that is, essentially, whenever \eqref{prediction first stage} and \eqref{prediction second stage} depart from the extensively researched and documented case of linear programming, stochastic programming resorts to approximate methods involving sampling. These methods originate from two broad areas of research and perspectives: \textit{Monte Carlo stochastic programming }\citep[e.g.,][]{DeMeBayr14, Shap03} and \textit{simulation optimization} \citep[e.g.,][]{Fu15}. In the former, the solution methods may leverage available knowledge about the inner structure of \eqref{prediction second stage}. In the latter, \eqref{prediction second stage} is viewed as a black box and the available knowledge consists solely in the ability to evaluate the solution ${\mathbf{y}}^*({\mathbf{x}}_{\text{a}}, {\mathbf{x}}_{\text{u}})$. (That is, ${\mathbf{y}}^*({\mathbf{x}}_{\text{a}}, {\mathbf{x}}_{\text{u}})$ may be computable with a standard OR solver without any assumption, for instance, about 
 closed-form derivatives with respect to ${\mathbf{x}}_{\text{a}}$ or ${\mathbf{x}}_{\text{u}}$.) An approximate solution to the problem jointly defined by \eqref{prediction first stage} and \eqref{prediction second stage} may be obtained through one of the methods available from Monte Carlo stochastic programming or simulation optimization \textit{for each particular value of}  ${\mathbf{x}}_{\text{a}}$.

Methods where sampling occurs once at the outset of the solution process to convert stochastic optimization into deterministic optimization and where sampling occurs throughout the optimization process are respectively said to involve \textit{batch} or \textit{external} sampling and \textit{sequential} or \textit{internal} sampling. Methods originating from the perspective of Monte Carlo stochastic programming that are in principle available to solve \eqref{prediction first stage} and \eqref{prediction second stage} include \textit{sample average approximation} (external) described, e.g., in  \cite{ShapEtAl09} and \cite{KimEtAl15} as well as versions of \textit{stochastic approximation} (internal) where a knowledge of the inner structure of \eqref{prediction second stage} is introduced \citep[p.~230]{ShapEtAl09}. Methods originating from the perspective of simulation optimization that are in principle available to solve \eqref{prediction first stage} and \eqref{prediction second stage} include \textit{response surface methods} (internal) \citep{Klei15}, \textit{stochastic search} (internal) \citep{Andr15, Zabi15, Hu15} and versions of \textit{stochastic approximation} (internal) where knowledge of \eqref{prediction second stage} is limited to the ability to perform evaluations \citep{ChauFu15}.

Our attention is directed to real-time or high-repetition applications requiring the computation of solutions to \eqref{prediction first stage} and \eqref{prediction second stage} at a high speed. Now, it is typically considerably more time-consuming to solve \eqref{prediction first stage} and \eqref{prediction second stage} on demand for a particular value of ${\mathbf{x}}_{\text{a}}$ through one of the existing methods available from approximate stochastic programming than it is to solve a single instance of the deterministic problem \eqref{prediction second stage}. As a result, a methodology that succeeds in achieving on-demand prediction times of smaller order than the time it takes to solve one instance of \eqref{prediction second stage} is highly desirable. The methodology based on ML that we propose satisfies this condition. For example, in the application presented in Section~\ref{sec:application}, the solution of a single easiest instance of the deterministic problem \eqref{prediction second stage} with a solver requires up to a minute whereas our methodology based on ML can yield predictions on demand within a millisecond.

\section{Methodology} \label{sec:methodology}
This section details the methodology outlined in Section~\ref{sec:intro}. Sections~\ref{sec:data} and~\ref{sec:mlp}  describe the generation of data and the ML approximation, respectively. Section~\ref{sec:aggregation} addresses the treatment of missing information through aggregation methods and the appropriate level of detail in TDOSs.

\subsection{Data Generation Process}
\label{sec:data}

The data used for ML derives from operational problem instances and their corresponding solutions. These may either result from controlled probabilistic sampling or, under restrictive conditions, may be collected from historical observations. In our context, controlled probabilistic sampling is advantageous because: (i) the selected sampling distribution is deemed an accurate representation of the stochasticity in the unknown features ${\textbf{x}}_{\text{u}}$ of the problem instances, (ii) it is possible to generate data at will according to a known sampling protocol and to evaluate the performance of ML training and model selection in any arbitrarily defined region in the feature space and (iii) it is possible to generate additional data for further training if the predictive performance is judged insufficient. The sampling distributions can be estimated from historical data. In contrast, the use of a dataset consisting of historical problem instances is only appropriate when attempting to mimic the behavior reflected in such data. Otherwise, sampling from historical data would likely introduce uncontrollable distortions in the resulting dataset. Indeed, observed instances result from censoring/constraining the space of admissible instances and stem from decision processes that should be accounted for but are often unknown in practice. In view of these limitations, we concentrate our attention on data generation through controlled probabilistic sampling.

The first step in the probabilistic data generation process is to sample a set of operational problem instances
$\{{\mathbf{x}}^{(i)}, ~i=1,\ldots,m\}$. Elements of ${\mathbf{x}}$ that are expected to vary in the intended application should be made to vary and covary in the dataset in commensurate ranges. We can generate problem instances through pseudo-random or quasi-random sampling. Data generation is meant to account for the actual uncertainty about the values of the elements of the input ${\mathbf{x}}$. In other words, the distributions from which we sample those values are viewed as describing this uncertainty and should be selected accordingly. Whereas simple pseudo-random sampling and stratified pseudo-random sampling are simple and easily applicable, it is also conceivable to apply alternative protocols in order to improve sample efficiency. For instance, importance sampling can artificially increase the abundance of data about infrequently observed characteristics of the problem instances. We refer to, for example, \cite{AsmuGlyn10} and \cite{Law14} for further details on data sampling and simulations.

We employ an existing solver to generate the operational solutions 
${\mathbf{y}}^*({\mathbf{x}}^{(i)})$, $i=1,\ldots,m$ to the problem instances. We note that the methodology does not require optimal solutions. The selection of the particular mechanism used in generating output labels for ML purposes depends on an assessment made by the subject matter expert (SME) managing the decision-making problem at hand (in our application, the SME would be an employee of the railway operator). The SME may choose to generate solutions from a number of alternative procedures: (i) use an exact solver and compute solutions up to a specified optimality gap (ii) rely on a heuristic solution method, or (iii) rely on human-constructed solutions. The SME selects the particular alternative presenting in his/her view the best trade off between closeness to perfect accuracy (achieved by optimal solutions) and resources invested in generating output labels for ML purposes.

In order to make efficient use of the computational resources available for ML, the operational, fully detailed optimal solutions ${\mathbf{y}}^*({\mathbf{x}}^{(i)})$, $i=1,\ldots,m$ are synthesized as 
$g({\mathbf{y}}^*({\mathbf{x}}^{(i)}))$, $i=1,\ldots,m$ according to a TDOS $g(.)$ whose level of detail accommodates without exceeding that required in the intended application. According to their level of detail, such descriptions vary in complexity. They may be highly structured and may feature a variable size. The complexity of the input vector ${\mathbf{x}}$ and the synthesized output vector $g({\mathbf{y}}^*({\mathbf{x}}))$, as well as the explicit constraints that may tie their elements impact the selection and performance of ML models and algorithms.

\subsection{ML Approximation}
\label{sec:mlp}

For clarity of exposition, we find it useful to disentangle at first the approximation of the TDOSs with ML from the treatment of the stochasticity in some of the inputs with ML. We therefore suppose for the moment that the input vector available at the time of prediction is equal to the full vector of input features, that is ${\textbf{x}} = {\textbf{x}}_{\text{a}}$ and ${\textbf{x}}_{\text{u}}$ is empty. Under this assumption, the aim of the ML approximation is simply to find the best possible prediction $\mathbf{y} = f({\mathbf{x}};\boldsymbol{\theta})$ of 
$g({\mathbf{y}}^*({\mathbf{x}}))$, where the approximator $f(\cdot;\cdot)$ is an ML model and $\boldsymbol{\theta}$ is a vector of parameters. The model $f(\cdot;\cdot)$ and $\boldsymbol{\theta}$ are selected through an ML algorithm, based on the available input-output data made up of $({\textbf{x}}, g({\mathbf{y}}^*({\mathbf{x}})))$
pairs. The models under consideration and the algorithms used in their training and selection must conform with the structure embodied in the input and output vectors ${\textbf{x}}$ and $g({\mathbf{y}}^*({\mathbf{x}}))$, and must also uphold the constraints that may explicitly relate their individual elements. The choice of a model and an algorithm necessarily depends on the exact application at hand and there is in ML a range of classification and regression approximators available for these purposes. The ML algorithm that we apply is standard and can be broadly summarized as follows:

 \begin{enumerate}
 \item{} The full dataset is divided at random between training, validation and test sets.
 \item{} Training and validation loss functions are selected.
 \item{} Parameters of candidate models are tuned through minimization of average training loss, i.e. empirical risk, over training set.
 \item{} Performances of trained candidate models are assessed and compared based on average validation loss, i.e. generalization error, measured over the validation set.
 \item{} Additional data is generated and processed if the performance on the validation set is unsatisfactory.
 \item{} The model achieving the lowest generalization error over the validation set is retained.
 \item{} The model performance is finally evaluated based on the average validation loss, measured over the test set.
 \item{} Provided the selected model demonstrates sufficient accuracy, it is used as a high speed, low marginal cost, on-demand predictor for the operational solution of any problem instance.
 \end{enumerate}

Additional challenges arise when the input vector available at the time of prediction is not equal to the full vector of input features, that is, whenever ${\textbf{x}} \neq {\textbf{x}}_{\text{a}}$. Those issues are addressed in the next section. However, the ML algorithm above remains essentially unchanged since the treatment of the stochasticity in ${\textbf{x}}_{\text{u}}$ with ML hinges on the particular definition of the  input-output data pairs that are supplied to ML.

\subsection{Aggregation and Subselection} \label{sec:aggregation}

The treatment of stochasticity in ${\textbf{x}}_{\text{u}}$ with ML can proceed in a number of ways using \emph{aggregation methods}. All ultimately translate into a particular definition for the input-output data pairs that are supplied to ML and all leverage the probabilistic information embodied in the joint distribution of ${\textbf{x}}_{\text{u}}$ conditional on the $\sigma$-algebra generated by ${\textbf{x}}_{\text{a}}$, say $\sigma({\textbf{x}}_{\text{a}})$, so as to ``aggregate'' the probability mass distributed over the support of ${\textbf{x}}_{\text{u}}$. We shall thus say that input features ${\textbf{x}}_{\text{u}}$ whose values are unavailable at the time of prediction (e.g., railcar capacities and container weights, in our application) are to be aggregated.

The more appealing aggregation methods are those involving the replacement of $\Phi_{{\mathbf{x}}_{\text{u}}} \lbrace \lVert .\rVert \mid {\mathbf{x}}_{\text{a}} \rbrace$ in \eqref{prediction first stage} with a sample version or a closed-form approximation (\emph{aggregation over outputs} for short) rather than the direct substitution of a $\sigma({\textbf{x}}_{\text{a}})$-measurable predictor for ${\textbf{x}}_{\text{u}}$ (\emph{aggregation over inputs} for short). 
Aggregation over inputs is simple but only acceptable when the distribution of ${\textbf{x}}_{\text{u}}$ has small variance. We focus on the treatment of stochasticity in ${\textbf{x}}_{\text{u}}$ with aggregation \emph{over outputs through ML approximation} in view of its simple application, low computational demand and, as we show in Section~\ref{sec:application}, good empirical performance. This method proceeds implicitly by supplying the dataset 
$\{({\mathbf{x}}^{(i)}_{\text{a}},
g({\mathbf{y}}^*({\mathbf{x}}^{(i)}_{\text{a}}, {\mathbf{x}}^{(i)}_{\text{u}})),~i=1,\ldots,m\}$ to ML. The prediction $\widehat{\mathbf{y}}^*({\mathbf{x}}_{\text{a}})$ of the TDOS $g({\mathbf{y}}^*({\mathbf{x}}_{\text{a}}, {\mathbf{x}}_{\text{u}}))$ is obtained from the trained model through $\widehat{\mathbf{y}}^*({\mathbf{x}}_{\text{a}}) \equiv f({\textbf{x}}_{\text{a}};\boldsymbol{\theta})$, where, again,  $f(\cdot;\cdot)$ is an ML model and $\boldsymbol{\theta}$ is a vector of parameters. The implementation of aggregation through ML approximation is straightforward since the required data is obtained directly from the sample of synthesized operational solutions.

We motivate in the following how the model $f({\mathbf{x}}_{\text{a}};\boldsymbol{\theta})$ resulting from aggregation through ML approximation can account for the stochasticity in ${\mathbf{x}}_{\text{u}}$. If a uniform law of large numbers holds so that the average validation loss converges stochastically towards the expectation of the validation loss with respect to the distribution of the variables that are sampled in the data \citep[e.g.,][]{Vapn99}, then we can argue that this aggregation method indeed minimizes an approximation to the expected validation loss with respect to ${\mathbf{x}}_{\text{a}}$ \emph{as well as} ${\mathbf{x}}_{\text{u}}$. In other words, ML can in this case be viewed as minimizing an approximation to the expected discrepancy between the exact TDOS 
$g({\mathbf{y}}^*({\mathbf{x}}_{\text{a}}, {\mathbf{x}}_{\text{u}}))$ resulting from knowledge of $[{\mathbf{x}}_{\text{a}}, {\mathbf{x}}_{\text{u}}]$ and the predictor $f({\mathbf{x}}_{\text{a}};\boldsymbol{\theta})$ based solely on the knowledge of ${\mathbf{x}}_{\text{a}}$. Furthermore, if $\Phi_{{\mathbf{x}}_{\text{u}}} \lbrace \lVert .\rVert \mid {\mathbf{x}}_{\text{a}} \rbrace$ stands for the expectation of a particular loss function and if the latter agrees with the loss function applied in ML validation, then we may argue that the method of aggregation through ML produces indeed a \emph{bona fide} approximator of $\bar{\mathbf{y}}^*({\mathbf{x}}_{\text{a}})$. The introduction of the expectation operator and the $L_2$-norm in $\Phi_{{\mathbf{x}}_{\text{u}}} \lbrace \lVert .\rVert \mid {\mathbf{x}}_{\text{a}} \rbrace$ is especially useful since the resulting ML approximator $\widehat{\mathbf{y}}^*({\mathbf{x}}_{\text{a}})$ may then be viewed as approximating the expectation of
$g({\mathbf{y}}^*({\mathbf{x}}_{\text{a}}, {\mathbf{x}}_{\text{u}}))$, conditionally on $\mathbf{x}_{\text{a}}$.

Selecting the scope of analysis, namely determining which variables of ${\textbf{x}}$ and domains thereof to include in the operational problem, the level of detail of the TDOS and the partition $[{\mathbf{x}}_{\text{a}}, {\mathbf{x}}_{\text{u}}]$ achieves a compromise between statistical precision, accuracy and tractability. Thus, it may be acceptable to exclude low probability regions in the support of certain relevant variables or even some relevant but judged-less-critical variables altogether from consideration in order to gain statistical precision at the expense of some accuracy. For short, we shall call such a restriction in the scope of analysis \emph{subselection}. For instance, in our application, it was judged useful to disregard infrequent railcar types and container lengths from the support of ${\textbf{x}} = [{\textbf{x}}_{\text{a}}, {\textbf{x}}_{\text{u}}]$ and the scope of analysis in order to gain
precision and tractability in ML. 

The excluded support and/or excluded variables and their complement constitute a partition of the whole support/set of variables that gives rise to an informational $\sigma$-algebra, say $\mathcal{G}$. The probabilistic knowledge that embodies these exclusions may be represented by the joint probability distribution of $[{\textbf{x}}_{\text{a}}, {\textbf{x}}_{\text{u}}]$ conditional on $\mathcal{G}$. If useful, for instance for the purpose of drawing formal comparisons between results with and without exclusions, conditioning with $\mathcal{G}$ may be introduced explicitly, possibly in addition to conditioning with $\sigma({\textbf{x}}_{\text{a}})$. Otherwise, the ML procedures can be blind to the exclusions. We proceed in this manner in our application.

\section{Application} \label{sec:application}
We use the double-stack intermodal railcar LPP  \citep{MantEtAl17} to illustrate the proposed methodology. The LPP can be briefly described as follows. Given a set of containers and a set of railcars, determine the subset of containers to load and the exact way of loading them on a subset of railcars. The objective consists in minimizing the total cost of containers left behind and partly filled railcars. The solution depends on individual characteristics of containers and railcars. Containers are characterized by their length, height, standardized type, content and weight. In North America, double-stack intermodal railcars comprise one to five platforms and each platform has a lower and an upper slot where containers may be loaded. Crucially, railcars are characterized by the weight capacity and tare of each platform and by the specific loading capabilities associated with their standardized type. We can express the loading capabilities with a set of loading patterns enumerating all possible ways in which containers of diverse lengths can be placed in the lower and upper slots of each platform. In general, the loading capabilities cannot be decomposed by platform, which leads to sets of loading patterns of high cardinality. \cite{MantEtAl17} propose an ILP formulation that can be solved in seconds or in minutes, depending on the size of the problem, using a commercial solver. 

\begin{figure}
\begin{center}
\includegraphics[width=0.8\textwidth]{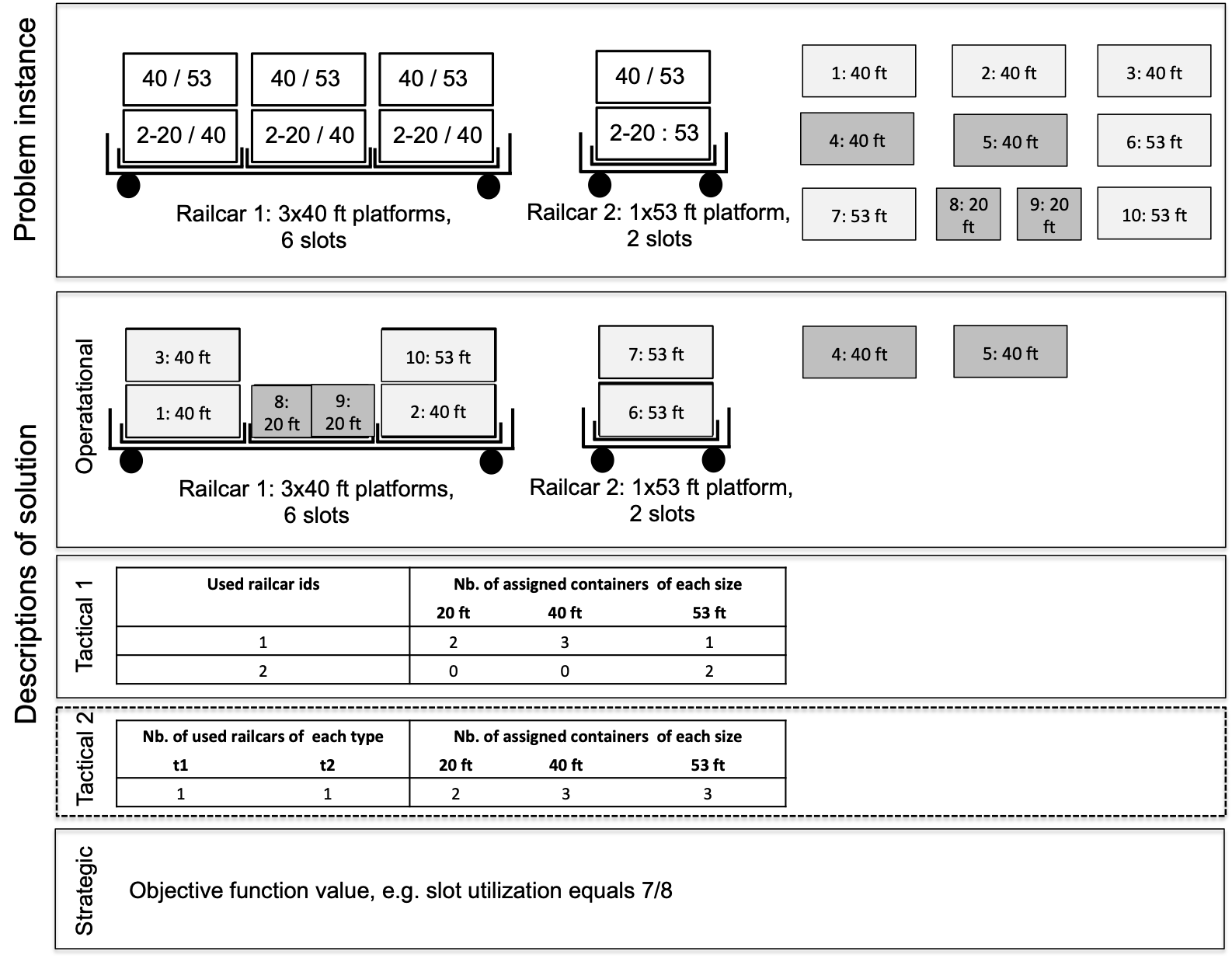}
\caption{Example of the double-stack railcar LPP}
\label{fig:loadplanning_example}
\end{center}
\end{figure}

Figure~\ref{fig:loadplanning_example} depicts a small example of a problem instance along with four descriptions of the optimal solution at different levels of detail. The instance contains ten containers of three different lengths: 20 feet (ft), 40~ft and 53~ft. For the sake of simplicity, we do not report exact weights but note that containers drawn in dark gray are considerably heavier than those drawn in light gray. The instance contains two railcars: one with three 40~ft platforms and another with one 53~ft platform. The numbers in each slot indicate the feasible assignments with respect to container sizes: Containers exceeding the length of the platform cannot be loaded in a bottom slot and 20~ft containers cannot be loaded in a top slot. 

The operational solution is illustrated immediately below the problem instance in Figure~\ref{fig:loadplanning_example}. The solution makes use of the two railcars and a subset of the containers are assigned to slots. In this example, one top slot is not used because of weight constraints and two heavy containers are not assigned. If the objective function is defined as the slot utilization, then its value is $7/8$. The latter corresponds to the strategic description of the operational solution, shown in the bottom part of the figure. Two alternative TDOSs whose levels of detail are intermediate are also depicted in the figure. One (labeled Tactical 1) specifies for each individual railcar the numbers of containers of each size that are assigned to it. The other, less detailed (labeled Tactical 2), specifies the numbers of railcars of each type that are used in the solution and the numbers of containers of each size that are assigned. There is a trade-off between the level of detail that is required and the difficulty of the prediction task. Thus, Tactical 1 and 2 naturally require output representations of variable and fixed length, respectively. Tactical 2 is used in our application.

We use our application to illustrate the notation: Feature vector ${\textbf{x}}$ contains the detailed information about the problem instance required to solve the LPP. Subvector ${\textbf{x}}_{\text{a}}$ reports features that are known at time of prediction: total numbers of available railcars of each type and of available containers of each length. Subvector ${\textbf{x}}_{\text{u}}$ reports features that are unknown at time of prediction: individual gross weights of containers. The problem described in \eqref{prediction second stage} corresponds to the ILP formulation of \cite{MantEtAl17}. The detailed assignment ${\mathbf{y}}^*({\mathbf{x}}_{\text{a}}, {\mathbf{x}}_{\text{u}})$ of containers to slots in the operational solution is illustrated in the second panel of Figure~\ref{fig:loadplanning_example}.
Tactical 1 and 2 constitute two examples of
$\bar{\mathbf{y}}^*({\mathbf{x}}_{\text{a}}, {\mathbf{x}}_{\text{u}}) : \equiv g({\mathbf{y}}^*({\mathbf{x}}_{\text{a}}, {\mathbf{x}}_{\text{u}}))$ that can be obtained from the detailed solution.

Our methodological proposal cannot be viewed as supplying heuristic solutions to the LPP. The LPP is a deterministic operational problem whose solution is computed after any uncertainty regarding the container weights has been resolved and specifies the finely detailed assignment of the containers to the available railcars. In contrast, we address the stochastic problem taking place at tactical standpoint when container weights are still unknown. We propose to approximate with ML the expectation of the TDOS. This expectation accounts for the uncertainty in container weights, is conditional upon available railcars and provides support for the tactical decisions.

\subsection{Subselection of Containers and Railcars and Aggregation} \label{sec:aggSubContsCars}

Container data may be transformed by subselecting lengths, heights, standardized types and contents and by aggregating weights. Similarly, railcar data may be transformed by subselecting standardized types and by aggregating weight capacities and tares. Hence, in order to ensure that the LPP remains manageable, container lengths, heights, standardized types, and contents have been subselected to retain only the 2 most relevant lengths, 40 ft and 53 ft, a single height, a single standardized type and a single content. Similarly, railcar types have been subselected to retain the 10 most numerous ones that amount to nearly 90\% of the North American fleet. This results in input-output vectors of size 12.

Exact weight capacities and tares of the railcars are unknown at the time of prediction. We account for this through aggregation. In relation to railcars, aggregation is straightforward because weight capacities and tares vary very little for each given standardized type. Hence, population estimates of median capacities and tares conditional on type are reasonable representative values. This amounts to aggregation over input values.

A key challenge in our application is related to the container weights that are unknown at the time of prediction. In contrast to the railcar weight capacities, container weights are highly variable, even conditionally upon the values of other container characteristics. We therefore perform aggregation over output values in view of its superior theoretical underpinnings compared to aggregation over input values (see Section \ref{sec:aggregation}).

\subsection{Data Generation}
\label{dataGeneration}
We partition the available data into four classes, as reported in Table~\ref{tab:dataClasses}. This partition facilitates experiments where models are trained and validated on simpler instances and tested on either simpler (A), harder (B, C) or hardest ones (D).

\begin{table}[h]
\begin{center}
\begin{footnotesize}
\begin{tabular}{|clcc|}
\hline
Class & Description & \# of containers & \# of platforms \\ \hline
A & Simpler ILP instances & [1, 150] & [1, 50] \\ 
B & More containers than A (excess demand) & [151, 300] & [1, 50] \\ 
C & More platforms than A (excess supply) & [1, 150] & [51, 100] \\
D & Largest and hardest instances & [151, 300] & [51, 100] \\
\hline
\end{tabular}
\caption{Data classes}
\label{tab:dataClasses}
\end{footnotesize}
\end{center}
\end{table}

We generated data by randomly sampling and distributing the total number of platforms among railcars belonging to the 10 standardized types, by randomly sampling the numbers of containers of each length and by randomly sampling the weight of each container given its length. In detail, sampling the container gross weights proceeded as follows based on historical data: First, we determined the empty/non-empty state of a container through a Bernouilli experiment where the probability that a container is empty conditionally upon its length has been estimated from container transportation data. Second, conditionally on the container not being empty, we sampled its net weight from a uniform distribution over values ranging between 10\% and 90\% of its net capacity given length. Third, we equated the generated total weight of a non empty container to the sum of the generated net weight given length and the \emph{a priori} estimate of median tare given length. Table~\ref{tab:dataGenerated} reports the number of examples for each data class. In order to analyze sample efficiency, we chose to generate a large number of instances (20M) of the easiest class A and we denote this dataset $A'$. For the sake of comparison, datasets $A''$-$D''$ all contain the same number of instances (200K). We randomly divided each dataset into training (64\%), validation (16\%) and test (20\%) sets.

Each generated instance of the LPP was solved with IBM ILOG CPLEX 12.6 down to an optimality gap of at most 5\%. The solutions in the resulting problem instance-solution pairs were described with a limited subset of features: numbers of railcars of each type and numbers of containers of each length used in the loading solution. The objective of the ILP formulation was set so as to enforce the following priorities in lexicographic order: maximize total number of containers loaded, minimize total length of railcars used, maximize total length of containers loaded. Table~\ref{tab:dataGenerated} reports the percentiles of computing times per instance using three out of the six cores of an Intel Xeon X5650 Westmere 2.67 GHz processor. For instance, the median time required to solve a deterministic instance of class A down to an optimality gap of at most 5\% is equal to 0.48~s, the median for class D is 5.44~s and we note a sizable variability as the 95th percentile is 1.67~s for class A and 20.89~s for class D.

\begin{table}[h]
\begin{center}
\begin{tabular}{|ccccc|}
\hline
Data & \# instances & \multicolumn{3}{c|}{Time percentiles (s)} \\
class/set & &  $P_5$ & $P_{50}$& $P_{95}$ \\ \hline
$A'$ & 20M & 0.007 & 0.48 & 1.67 \\
$A''$ & 200K & 0.011 & 0.64 & 2.87 \\ \hline
$B''$ & 200K & 0.02 & 1.26 & 3.43 \\
$C''$ & 200K & 0.72 & 2.59 & 6.03 \\ \hline
$D''$ & 200K & 2.64 & 5.44 & 20.89 \\
\hline
\end{tabular}
\caption{Data generation}
\label{tab:dataGenerated}
\end{center}
\end{table}

\subsection{Measuring the Predictive Performance} \label{meas_pred_perf}

We summarize and compare the predictive performances based on \eqref{eq:MAE}, the sum of (i) mean absolute prediction error (MAE) measured over the number of used slots \eqref{eq:MAEslots} and of (ii) mean absolute prediction error measured over the number of loaded containers \eqref{eq:MAEcont} in output solution. Thus measuring the prediction errors in terms of the $L_1-$ instead of the $L_2$-norm will facilitate the interpretation of the predictive performance. The three criteria are defined as follows:

\begin{align} 
\label{eq:MAE}
\mathit{MAE}=\displaystyle \frac{1}{m} \sum_{i=1}^{m}\sum_{j=1}^{12} | \widehat{y}_j^{(i)} - y_j^{(i)}| s_j, \\ 
\label{eq:MAEslots}
\mathit{MAE}_{\mathit{slots}}=\displaystyle \frac{1}{m} \sum_{i=1}^{m}\sum_{j=1}^{10} | \widehat{y}_j^{(i)} - y_j^{(i)}| s_j, \\ 
\label{eq:MAEcont}
\mathit{MAE}_{\mathit{conts}}=\displaystyle \frac{1}{m} \sum_{i=1}^{m}\sum_{j=11}^{12} | \widehat{y}_j^{(i)} - y_j^{(i)}|, 
\end{align}
where $m$ is the number of examples and $s_j, ~j=1,\ldots,10$, equals the number of slots on railcar type $j$. Notice that $s_{11}=s_{12}=1$ do not appear in (\ref{eq:MAEcont}). To draw a more complete picture of the predictive performance, we also calculate empirical quantiles of the set of absolute errors $\{AE^{(i)},~i=1,\ldots,m\}$,  
where the absolute error $AE^{(i)}$ associated with observation $i$ is given by
\begin{align} 
\label{eq:AE}
AE^{(i)}=\displaystyle \sum_{j=1}^{12} | \widehat{y}_j^{(i)} - y_j^{(i)}| s_j. 
\end{align}

\subsection{A Lower Bound from Stochastic Programming} \label{sec:stochProgSolution}

The stochastic limits of statistic \eqref{eq:MAE} that are estimated by ML for various models, learning algorithms and data sets are all bounded from below and away from zero due to the stochastic nature of the prediction problem. Such lower bounds
make it possible to assess how well the ML approximators fare and if attempts at improving their performance by increasing their capacity and/or the size of training data are worthwhile.
Estimates of the relevant lower bounds can be calculated from stochastically consistent solutions to a particular implementation of the optimal prediction problem \eqref{prediction first stage} and \eqref{prediction second stage}. 

In order (i) to calculate an approximate lower bound for the prediction error achievable by the ML approximators and (ii) to illustrate the parallel between the proposed methodology based on ML and the alternatives offered by approximate stochastic programming, we detail the application of one particular method in the context of the LPP. We select the method of SAA \cite[see, e.g.,][p.~155]{KimEtAl15,ShapEtAl09} in view of its position as a \emph{de facto} standard in approximate stochastic programming: it is broadly applicable, commonly used, based on simple principles and supported by an abundant knowledge of its properties, both theoretical and empirical. We stress that an application of approximate stochastic programming to \eqref{prediction first stage} and \eqref{prediction second stage} seeks a solution for each particular value of $\textbf{x}_{\text{a}}$. Hence, unless the domain of $\textbf{x}_{\text{a}}$ is both finite and small, the solutions must be computed on demand, one at a time. In contrast, our proposed methodology outputs a prediction function defined over the domain of $\textbf{x}_{\text{a}}$. This function is computed in advance with ML and used later on to yield solutions on demand.

The LPP specification of the general optimal prediction problem of \eqref{prediction first stage} and \eqref{prediction second stage} is 
 \begin{equation}
\bar{\mathbf{y}}^*({\mathbf{x}}_{\text{a}}) : \equiv \arg \inf_ {\bar{\mathbf{y}}({\mathbf{x}}_{\text{a}}) \in \bar{\mathcal{Y}}({\mathbf{x}}_{\text{a}})}
E_{{\mathbf{x}}_{\text{u}}} \lbrace \sum_{j=1}^{12} \lvert
\bar{\mathbf{y}}_j({\mathbf{x}}_{\text{a}}) - g_j({\mathbf{y}}^*({\mathbf{x}}_{\text{a}}, {\mathbf{x}}_{\text{u}})) \rvert s_j \mid {\mathbf{x}}_{\text{a}} \rbrace
\label{eq:prediction_first_stage_LPP}
\end{equation}

\begin{equation}
{\mathbf{y}}^*({\mathbf{x}}_{\text{a}}, {\mathbf{x}}_{\text{u}}) :\equiv \arg \inf_{{\mathbf{y}} \in \mathcal{Y}({\mathbf{x}}_{\text{a}}, {\mathbf{x}}_{\text{u}})} C({\mathbf{x}}_{\text{a}}, {\mathbf{x}}_{\text{u}}, {\mathbf{y}})
\label{eq:prediction_second_stage_LPP}
\end{equation}
and the application of the SAA method proceeds through the following steps:

\begin{enumerate}
\item{} Fix a particular value for $\textbf{x}_{\text{a}}$.
\item{} Select a sampling distribution for $\textbf{x}_{\text{u}}$ and generate a sample of size $N$ $\{\textbf{x}_{\text{u}}^{(i)},~i=1,\ldots,N\}$.
\item{} Define sample versions for \eqref{eq:prediction_first_stage_LPP} and \eqref{eq:prediction_second_stage_LPP} as
\begin{equation}
\bar{\mathbf{y}}^*({\mathbf{x}}_{\text{a}}) : \equiv \arg \inf_ {\bar{\mathbf{y}}({\mathbf{x}}_{\text{a}}) \in \bar{\mathcal{Y}}({\mathbf{x}}_{\text{a}})}
\sum_{i=1}^{N} \sum_{j=1}^{12} \lvert
\bar{\mathbf{y}}_j({\mathbf{x}}_{\text{a}}) - g_j({\mathbf{y}}^{*(i)}({\mathbf{x}}_{\text{a}}, {\mathbf{x}}^{(i)}_{\text{u}})) \rvert s_j
\label{eq:prediction_first_stage_LPP_sample}
\end{equation}
\begin{equation}
{\mathbf{y}}^{*(i)}({\mathbf{x}}_{\text{a}}, {\mathbf{x}}^{(i)}_{\text{u}}) :\equiv \arg \inf_{{\mathbf{y}}^{(i)} \in \mathcal{Y}({\mathbf{x}}_{\text{a}}, {\mathbf{x}}^{(i)}_{\text{u}})} C({\mathbf{x}}_{\text{a}}, {\mathbf{x}}^{(i)}_{\text{u}}, {\mathbf{y}}^{(i)}), ~i=1,\ldots,N
\label{eq:prediction_second_stage_LPP_sample}
\end{equation}

\item{} Solve the $N$ operational problems in \eqref{eq:prediction_second_stage_LPP_sample}.
\item{} Solve the ILP \eqref{eq:prediction_first_stage_LPP_sample}. 
\end{enumerate}

The fact that the set of solutions to \eqref{eq:prediction_second_stage_LPP_sample} is finite for a given value of $\textbf{x}_{\text{a}}$ simplifies the analysis of the statistical properties of the SAA method. Hence, the distance between solution(s) to \eqref{eq:prediction_first_stage_LPP_sample} and solution(s) to \eqref{eq:prediction_first_stage_LPP} is shown to converge strongly to zero with respect to the so-called ``number of scenarios'' $N$ \citep{KleyEtAl02}. Although some theoretical bounds are available, determining with sufficient precision the speed of this convergence and an appropriate value for $N$ is essentially a problem- and data-specific issue that must be resolved empirically. 
Despite the lack of precise advance knowledge about the appropriate value for $N$, we realize at once that the application of SAA on demand for each value of $\textbf{x}_{\text{a}}$ is too costly for our computing time budget that is less than the time it takes to solve the deterministic problem. For this reason, we do not use state-of-the-art solution algorithms for the SAA problem. Instead, we compute solutions using the default setting of ILOG CPLEX 12.6. We report computing times for the sake of completeness (last column in Table~\ref{tab:SAA_class_D}).

In an attempt to infer an approximate lower bound, we calculated SAA solutions for $N \in \lbrace 5, 10, 25, 50, 75, 99 \rbrace$ through these steps:
\begin{enumerate}
    \item Generate a dataset of the target class (A, B, C or D) from the sampling distributions described in Section~\ref{dataGeneration} through a two-stage sampling process. In the first stage, randomly sample 100K values for $\textbf{x}_{\text{a}}$. In the second stage, for each one of the first stage values, randomly sample 100 values for $\textbf{x}_{\text{u}}$.
    \item For each one of the 100K values of $\textbf{x}_{\text{a}}$ resulting from the first stage, compute the 100 deterministic LPP solutions corresponding to this value of $\textbf{x}_{\text{a}}$ and to each one of the 100 values of $\textbf{x}_{\text{u}}$, respectively. Compute the SAA solution with the first $N$ out of the 100 load planning solutions, as in \eqref{eq:prediction_first_stage_LPP_sample} and \eqref{eq:prediction_second_stage_LPP_sample}. Compute the absolute error between the SAA solution and the 100th deterministic load planning solution as well as the total time required to compute the SAA solution.
    \item Compute the empirical distributions of the absolute errors and on-demand computing times incurred over the 100K repetitions resulting from the 100K values of $\textbf{x}_{\text{a}}$ sampled in the first stage.
\end{enumerate}

We start by analysing results for the hardest class of instances (D) reported in Table~\ref{tab:SAA_class_D}. The sample MAE values associated with the SAA solutions monotonically decrease from 2.879 to 2.682 at a decreasing rate as a function of the number of scenarios and are nearly identical for numbers of scenarios equal to 75 and 99. Hence, it is reasonable to view 2.682 as being in the vicinity of the lower bound for the optimal prediction problem over data class D. From Chebyshev's inequality \citep[e.g.,][]{KendEtAl1987}, we know that the probability that the exact lower bound lies approximately inside three times the estimated standard error of estimate 8.08E-02 from either side of the sample MAE 2.682 is greater than 89\%.
Therefore, we define a confidence interval around this value as $[2.4396, 2.9244]$, which we refer to as the \emph{SAA lower bound over data class D}. In a similar fashion, we compute SAA lower bounds over data classes A $[0.7978, 0.8502]$, B $[0.7584, 0.8056]$ and C $[3.0593, 3.1787]$. 

\begin{table}[!htbp]
    \centering
    \begin{tabular}{|l|cc|cc|}
    \hline
    Nb. of scenarios$^{*}$ & \multicolumn{2}{c|}{MAE}  & \multicolumn{2}{c|}{Computing time [s]} \\ 
        & Est. mean & Est. std dev & Est. mean & Est. std dev \\\hline
    5 & 2.879 & 26.412 & 42.144 & 49.748 \\
      & (8.35E-02) & & (1.57E-01) & \\\hline
    10 & 2.753 & 25.731 & 84.070 & 89.502 \\
      & (8.14E-02) & & (2.83E-01) & \\\hline
    25 & 2.691 & 25.559 & 209.776 & 207.328 \\
      & (8.08E-02) & & (6.56E-01) & \\\hline
    50 & 2.687 & 25.568 & 419.700 & 403.982 \\
      & (8.09E-02) & & (1.28E-00) & \\\hline
    75 & 2.681 & 25.535 & 628.900 & 599.345 \\
      & (8.08E-02) & & (1.90E-00) & \\\hline
    99 & 2.682 & 25.561 & 829.956 & 787.748 \\
      & (8.08E-02) & & (2.49E-00) & \\\hline
     \multicolumn{4}{l}{\small * number of sets of weights drawn for each example}\\
      \multicolumn{4}{l}{\small Standard error of estimate is reported between parentheses.}\\
      \end{tabular}
       \caption{Properties of the SAA predictor over class D}
       \label{tab:SAA_class_D}
    
\end{table}

\subsection{ML Approximation} \label{sec:mlapprox}
The predictive models are based on feedforward neural networks, a.k.a. multilayer perceptrons (MLP). From their introduction several decades ago until recently,  MLPs had demonstrated modest success in ML. However, through the recent algorithmic advances that have occurred in the subfield of ML known as deep learning, they have become simple but powerful generic approximators. They are useful in real applications whenever input and output vectors have short fixed lengths and do not feature complex structures (e.g., images and sound typically require other models).

The mapping between inputs and outputs could be interpreted as a classification or as a regression problem and we implemented the two corresponding architectures. The resulting families of models are hereafter named ClassMLP and RegMLP, respectively. Both families feature 12 units in their input layer (one integer unit for each railcar type and for each container length) and rectified linear units (ReLU) are used as activation functions in their hidden layers. The two families differ with respect to their output layer, the manner in which input-output constraints are upheld and the loss function used in their training.

On the one hand, ClassMLP outputs 12 discrete probability distributions (one for each railcar type and for each container length) that are each modeled with a softmax operator. The supports of these distributions are the sets of possible numbers of railcar of each type and of containers of each length. Thus, concatenation of the 12 distributions yields an output layer of size 812 when the numbers of railcars platforms of each type and of containers of each length used in the solution may respectively vary from 1 to 50 and from 1 to 150. The following constraints are enforced at training, validation and testing times: for each type of railcar and length of container, the number in output cannot exceed the number in input. This is done by computing the softmax over admissible outputs only. Training is conducted through likelihood maximization where we treat output distributions as independent. 

On the other hand, RegMLP outputs 12 scalars that are rounded to the nearest integer, input-output inequalities are enforced only at validation and testing times and training is conducted through minimization of the sum of absolute errors incurred in predicting output numbers for railcars of each type and containers of each length.
For both families, the assumption that outputs are conditionally mutually independent given inputs is implicit in their architecture.

Training of both ClassMLP and RegMLP was performed with mini-batch stochastic gradient descent and the learning rate adaptation was governed by the Adam (adaptive moment estimation) method \citep{KingBa14}. Regularization was ensured by early stopping. Hyperparameter selection included number and width of hidden layers as well as $L_1$ and $L_2$ regularization terms coefficients. Sets of hyperparameter values were generated randomly and the preferred set was determined according to validation results \citep{BergBeng12}. The ranges of values considered were as follows: [3, 13] for the number of hidden layers, [300, 1000] for the number of units per hidden layer, [0, 0.001] for the $L_1$ and $L_2$ regularization coefficients. For each ML model and each dataset, we tried 50 hyperparameter combinations in a random search and retained the best performing models.

Since the optimal prediction problem we face is new, we cannot compare our results to an existing benchmark. In order to assess the quality of the predictions, we compare the performance against a lower bound from stochastic programming, as introduced in Section~\ref{sec:stochProgSolution}. We also assess if the prediction task is easy, or even trivial. For this purpose, we present additional results obtained with, first, simple greedy algorithms and, second, simple prediction models, namely 
logistic regression (ClassMLP without hidden layers) and linear regression (RegMLP without hidden layers). Algorithm \ref{algo:HeurV} (HeurV) greedily accounts for the total number of slots available on each railcar but disregards all other constraints pertaining to the loading problem. In contrast, Algorithm \ref{algo:HeurS} (HeurS) greedily accounts for all constraints relevant to the loading problem, namely: (i) 53 ft containers can only be assigned to 53 ft slots whereas 40 ft containers can be assigned to any slot, (ii) for some railcars, some 53 ft top slots are only available  provided that 40 ft containers are loaded in the bottom slot. This algorithm also attempts to account for the lexicographic objective.

\begin{algorithm}
\caption{Very simple greedy heuristic (HeurV)}
\label{algo:HeurV}
\begin{algorithmic}
\WHILE{unassigned cont. and avail. car}
\FORALL{car type in car types}
\FORALL{car matching car type}
\STATE assign avail. cont(s) to car, alternating if possible between 40' and 53' cont(s);
\ENDFOR
\ENDFOR
\ENDWHILE
\end{algorithmic}
\end{algorithm}

\begin{algorithm}
\caption{Simple greedy heuristic (HeurS)}
\label{algo:HeurS}
\begin{algorithmic}
\WHILE{unassigned 53' cont. and avail. car with usable 53' slot(s)}
\STATE choose shortest among cars with greatest number of usable 53' slots not exceeding number of 53' conts still to assign;
\STATE otherwise, choose shortest among cars with smallest number of usable 53' slots;
\STATE assign as many 53' conts as possible to usable 53' slots on car;
\STATE assign as many 40' conts as possible to remaining available slots of car;
\ENDWHILE
\WHILE{unassigned 40' cont. and avail. car}
\STATE choose shortest among cars with greatest number of slots not exceeding number of 40' conts still to assign;
\STATE otherwise, choose shortest among cars with smallest number of slots;
\STATE assign as many 40' conts as possible to available slots of car;
\ENDWHILE
\end{algorithmic}
\end{algorithm}

\subsection{Results} \label{sec:results}

Table \ref{table:testingInsideOThrML_MAE} reports the sample MAE (\ref{eq:MAE}) incurred by each model over an independent test dataset similar to that used for training and validation. The aggregation of the unknown container weights is performed with the method of \emph{aggregation over output values through ML}, as discussed in Section~\ref{sec:aggregation}. Standard errors of the estimates are shown in parentheses.  We report results for models trained, validated and tested based on 200K i.i.d.\ examples of class A (dataset 200K-A, second column in the table), 20M i.i.d.\ examples of class A (dataset 20M-A, third column in the table) and 600K i.i.d examples made up of the union of 200K examples from each of the A, B and C classes (dataset 600K-ABC, fourth column in the table). Note that each figure reported for ClassMLP or for RegMLP corresponds to the most favorable set of hyperparameters according to the validation process (found through the random search described in Section~\ref{sec:mlapprox}).

A number of findings immediately emerge from the examination of Table~\ref{table:testingInsideOThrML_MAE}. For both datasets 200K-A and 20M-A comprising only class A examples (columns two and three), the average performances of both feedforward neural network models -- ClassMLP and RegMLP -- are very good in comparison with the SAA lower bound over data class A whose value is estimated in Section~\ref{sec:stochProgSolution} at $[0.7978, 0.8502]$. For example, the sample MAE achieved by ClassMLP and RegMLP over the 4M class A testing examples of dataset 20M-A are respectively equal to 0.965 and 0.985, when trained over the 12.8M class A training examples of dataset 20M-A and validated over the the 3.2M class A validation examples of dataset 20M-A. For the joint dataset 600K-ABC (column four) comprising the harder examples of classes B and C in addition to examples of class A, the average performances of ClassMLP and RegMLP also appear to be very good. For example, sample MAE for RegMLP modestly increases from 1.304 (dataset 200K-A, second column) to 2.109 (dataset 600K-ABC, fourth column). 
For all three datasets (columns two to four), the performances of ClassMLP and RegMLP are considerably better than those of the simple predictors (logistic regression, linear regression and the two heuristics) indicating that the task is not trivial.

We also note the following: First, the marginal value of using 100 times more training and validation examples is fairly small. For example, the sample MAE of RegMLP only increases from 0.985, when training and validating over the 12M training examples and 3.2M validation examples of 20M-A, to 1.304 when training and validating over the 120K training examples and 32K validation examples of 200K-A. Second, RegMLP features a slightly better average performance than ClassMLP, except for the 20M-A data. A possible explanation for this is that the pseudo-likelihood function used as a surrogate for MAE in training ClassMLP does not account for the magnitude of the prediction errors. 

The very good predictive performance of ClassMLP and RegMLP is confirmed by the examination of the distribution of absolute errors in Table~\ref{table:testingInsideOThrML_AE}: For example, at least 95\% of the absolute errors made by ClassMLP and RegMLP are smaller than or equal to 4. This is in stark contrast with the performances of the simple predictors whose distributions of absolute errors are highly skewed and whose median absolute error is either equal to 4 in the most favorable case (LogReg) or well beyond this figure elsewhere (LinReg, HeurV, HeurS).

\begin{table}[!htbp]
%\begin{scriptsize}
\begin{center}
\begin{tabular}{|c |c |c | c|}
\hline
%\multicolumn{1}{c|}{}& \bf A\_2S\_thr\_200K & \bf A\_1S\_thr\_20M & \bf ABC\_2S\_thr\_600K \\
\bf Data & \bf 200K-A & \bf 20M-A & \bf 600K-ABC \\
\bf \# examples & 200K & 20M & 600K \\
\hline \hline
\multirow{2}{*}{\bf ClassMLP} & 1.481 & 0.965 & 2.312  \\% exp12 xx, exp16 xx, exp13 xx (svp préserver ces commentaires)
 & {(0.018) } & {(0.002)} & {(0.014)}  \\
\cline{2-4}
\multirow{2}{*}{\bf LogReg} &  5.956 &  5.887 &  9.051 \\ % SL: A_1S_thr_20M ne semble pas avoir convergé encore tout à fait. 
 & {(0.029)} & {(0.003)} & {(0.027)}  \\
\hline
\multirow{2}{*}{\bf RegMLP} & 1.304 & 0.985 & 2.109 \\ % exp3 xx, exp16 xx, exp16 xx(svp préserver ces commentaires)
 & {(0.017)} & {(0.002) } & {(0.014)}  \\
\cline{2-4}
\multirow{2}{*}{\bf LinReg} & 18.306 & 18.372 &  39.907  \\
 & {(0.094)} & {(0.009) } & {(0.084)}   \\
\hline
\multirow{2}{*}{\bf HeurV} & 14.733 & 14.753 & 27.24  \\
 & {(0.075) } & {(0.008) } & {(0.083) }  \\
\cline{2-4}
\multirow{2}{*}{\bf HeurS} & 17.841 & 17.842 & 31.448 \\
 & {(0.083)} & {(0.008)} & {(0.089)}  \\
\hline
\multicolumn{4}{l}{\small Standard error of estimate is reported between parentheses.} \\
\end{tabular}
\caption{Testing over data similar to that used in training-validation: sample mean absolute errors (MAE)}
\label{table:testingInsideOThrML_MAE}
\end{center}
%\end{scriptsize}
\end{table}

\begin{table}[!htbp]
%\begin{scriptsize}
\begin{center}
\begin{tabular}{|c |c c c  c c c c  c|}
\hline
\bf Data & \multicolumn{8}{c|}{\bf 200K-A} \\
\bf \# examples & \multicolumn{8}{c|}{200K} \\
 \bf Percentiles & $P_{50}$ & $P_{60}$ & $P_{70}$ & $P_{80}$ & $P_{85}$ & $P_{90}$ & $P_{95}$ & $P_{99}$ \\
\hline \hline
\bf ClassMLP & 0 & 0 & 0 & 1 & 2 & 4 & 4 & 18 \\  
\cline{2-9}
\bf LogReg & 4 & 6 & 7 & 10 & 12 & 14 & 17 & 26 \\  
\hline

\bf RegMLP & 0 & 0 & 0 & 1 & 2 & 4 & 4 & 18 \\
\cline{2-9}
\bf LinReg & 11 & 14 & 19 & 30 & 38 & 47 & 61 & 82 \\  
\hline
\bf HeurV & 10 & 12 & 16 & 24 & 30 & 36 & 46 & 68 \\ 
\cline{2-9}
\bf HeurS & 13 & 17 & 23 & 31 & 35 & 41 & 52 & 72 \\
\hline
\end{tabular}
\caption{Testing over data similar to that used in training-validation: distribution of absolute errors}
\label{table:testingInsideOThrML_AE}
\end{center}
%\end{scriptsize}
\end{table}

\begin{table}[!htbp]
%\begin{scriptsize}
\begin{center}
\begin{tabular}{|c |c c c |c c  c|}
\hline
%\multicolumn{1}{c|}{}& \bf A\_2S\_bef\_200K &  \bf ABC\_2S\_bef\_600K \\
\bf Data & \multicolumn{3}{c|}{\bf 200K-A} \\
\bf \# examples & \multicolumn{3}{c|}{200K} \\
 \bf Percentiles & $P_5$ & $P_{50}$ & $P_{95}$ \\
\hline \hline
\bf ClassMLP & 2.6 & 2.9 & 3.2 \\ % exp9 xx, exp9 xx 
\hline
\bf RegMLP & 0.7 & 0.8 & 1.0 \\ % exp1 xx, exp13 xx(svp pr√©server ces commentaires) 
\hline
\bf HeurV & 0.3 & 0.4 & 0.8 \\ 
\bf HeurS & 0.3 & 0.7 & 1.6  \\ %
% \bf ClassMLP & \num{2.6e-3} & \num{2.9e-3} & \num{3.2e-3} & \num{2.3e-3} & \num{2.4e-3} & \num{2.5e-3} \\ % exp9 xx, exp9 xx 
% \hline
% \bf RegMLP & \num{7.1e-4} & \num{8.3e-4} & \num{1.0e-3} & \num{5.4e-4} & \num{6.3e-4} & \num{7.8e-4} \\ % exp1 xx, exp13 xx(svp préserver ces commentaires) 
% \hline
% \bf HeurV & \num{2.8e-4} & \num{4.2e-4} & \num{8.1e-4} & \num{2.8e-4} & \num{4.3e-4} & \num{8.0e-4} \\ 
% \bf HeurS & \num{2.7e-4} & \num{7.0e-4} & \num{1.6e-3} & \num{2.7e-4} & \num{6.9e-4} & \num{1.5e-3}  \\ %
\hline
\end{tabular}%timesT
\caption{Prediction time per instance (milliseconds)}
\label{table:timesTestingInside}
\end{center}
%\end{scriptsize}
\end{table}

\begin{figure}[!htbp]
\includegraphics[width=1.1\textwidth]{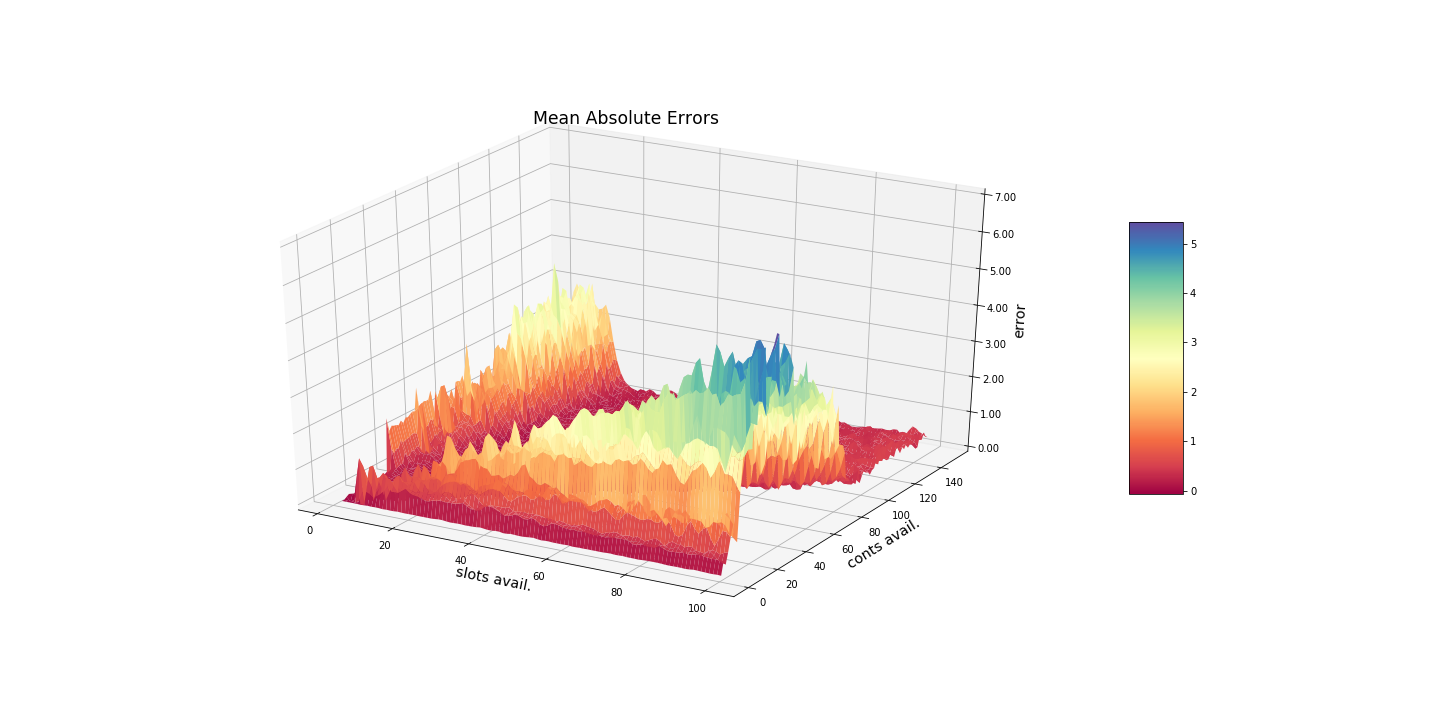}
\caption{MAE over instances with specified numbers of available slots and containers, RegMLP model and 20M-A data.}
\label{fig:errorsRegMLPA1SImplicit20MSize2010}
\end{figure}

Figure~\ref{fig:errorsRegMLPA1SImplicit20MSize2010} displays the MAE in relation to the number of available slots and the number of available containers (input) for the RegMLP model and 20M-A data. It shows that errors occur mainly in conditions of excess supply or excess demand.

Table \ref{table:timesTestingInside} provides information on the distribution of the time required to compute a prediction based on input data similar to that used for training and validation of the predictor. For example, the median time required to compute a prediction based on model RegMLP when input belongs to class A is 0.8 milliseconds. 
As clearly shown by the closeness of the 5th, 50th and 95th percentiles, the distribution of the prediction time is highly concentrated and we ought to expect small variations among computing times around the median value. Furthermore, it is expected that the figures of Table \ref{table:timesTestingInside} will vary little across input classes with a similar model. Computational speed should instead depend on model complexity (in our case number and width of hidden layers).

In comparison, Table~\ref{tab:dataGenerated} indicates that even if the prediction problem defined by \eqref{prediction first stage} and \eqref{prediction second stage} were deterministic, that is, if ${\textbf{x}} = {\textbf{x}}_{\text{a}}$ held, whence the solution of the operational load planning problem could be calculated in advance exactly, one would still face a highly dispersed computation time with a median ranging from 0.48 to 5.44 seconds according to the exact class of the input.

\paragraph{Extraneous Errors.}
In view of the higher costs of generating harder instances (i.e., solving harder problems), it is desirable that models that are trained and validated on simpler instances generalize to harder instances without specific training and validation. In contrast with the previous results where testing was conducted on data similar to that used for training and validation, we now focus on testing the performance over a set of class D data containing the hardest instances. We emphasize that instances of this nature have not been used for training-validation.

Table~\ref{table:testingOutsideOThrML} reports sample MAE achieved over the 200K class D examples of dataset 200K-D by the exact models whose performances over testing data of class A are reported in Table~\ref{table:testingInsideOThrML_MAE}. Standard deviations of the estimates are shown in parentheses and we report between brackets the range in performance achieved over all hyperparameter sets. Since the classification models cannot generate solution predictions for problem instances whose ranges of numbers of available containers and available railcars exceed those encountered during training and validation, we do not report any MAE values for the predictors ClassMLP and LogReg that are trained and validated over the class A examples of dataset 20M-A (Table~\ref{table:testingOutsideOThrML}, second column).

The following findings emerge. First, the performance of RegMLP that has been trained and validated on class A examples is still good when tested over class D examples. Compared to the SAA lower bound over data class D whose value is estimated in Section~\ref{sec:stochProgSolution} at $[2.4396, 2.9244]$, RegMLP achieves a sample MAE equal to 4.412 over the 200K testing examples of class D of dataset 200K-D (Table~\ref{table:testingOutsideOThrML}, second column), when trained over the 12.8M training examples of class A of dataset 20M-A and validated over the 3.2M validation examples of dataset 20M-A. Second, the testing performance over class D clearly benefits from performing training and validation over classes B and C in addition to class A. Thus, when trained over the 360K training examples of classes A, B and C of dataset 600K-ABC and validated over the 96K validation examples of classes A, B and C of dataset 600K-ABC, RegMLP achieves over the 200K testing examples of class D of dataset 200K-D a sample MAE equal to 2.372 (Table~\ref{table:testingOutsideOThrML}, second column) instead of 4.412. (Notice that confidence intervals can be calculated around the point estimates of MAE appearing in Table~\ref{table:testingOutsideOThrML}.)

The sample MAE values reported between brackets indicate that the range of the performances achieved on D over all hyperparameter sets considered at validation is wide. For example, it varies between 2.481 and 24.702 for RegMLP when training and validating over dataset 20M-A and testing over dataset 200K-D. We note that some hyperparameter sets achieving close to best validation performance perform poorly on D. The range of performances achieved over all hyperparameter sets considered at validation is reduced when training and validating on B and C in addition to A (last column). 

This opens up for questions concerning alternative data generation procedures. For example, less expensive labeling of the hardest instances could be accomplished by setting the maximum optimality gap to a more lenient value or even use solutions obtained with a heuristic.
Training and validation could then be performed on these instances as well.

Finally, Table~\ref{table:timesTestingOutside} provides information on the distribution of the time required to compute a prediction based on extraneous data. The remarks made in relation to Table~\ref{table:timesTestingInside} are still valid and the figures reported here are not markedly different. 

\begin{table}[htbp!]
%\begin{scriptsize}
\begin{center}
\begin{tabular}[]{|c|c|c|}
\hline
%\multicolumn{1}{c|}{}& \bf A\_1S\_thr\_20M & \bf ABC\_2S\_thr\_600K \\
\bf Training-validation data & \bf 20M-A & \bf 600K-ABC \\ \hline
\bf Testing data & \bf 200K-D & \bf 200K-D \\
\hline \hline
\multirow{2}{*}{\bf ClassMLP} & NA & 14.831 [13.161, 23.892] \\ % exp16 xx, exp13 xx
 &  & {(0.072) } \\
\cline{2-3}
 \multirow{2}{*}{\bf LogReg} & NA & 29.568 \\  %MAE: 29.5682 MAE_std: 0.0654
 &  & {(0.065)} \\
\hline
 \multirow{2}{*}{\bf RegMLP} & 4.412 [2.481, 24.702] & 2.372 [2.355, 3.305]  \\ % exp16 xx, exp16 xx
& {(0.050) } & {(0.051) }  \\
\cline{2-3}
\multirow{2}{*}{\bf LinReg} & 24.560 & 72.847 \\
 & {(0.064) } & {(0.060) }  \\
\hline
 \multirow{2}{*}{\bf HeurV} & 33.737 & 33.737 \\ % Both are equal since it's the same algorithm applied to the same dataset(d_implicit)
 & {(0.085) } &  {(0.085) }  \\
\cline{2-3}
\multirow{2}{*}{\bf HeurS} & 43.303 & 43.303 \\ 
& {(0.089) } &  {(0.089) }  \\
\hline
\end{tabular}
\caption{Testing on class D instances (not used for training-validation): sample mean absolute errors (MAE)}
\label{table:testingOutsideOThrML}
\end{center}
%\end{scriptsize}
\end{table}

\begin{table}[!htbp]
%\begin{scriptsize}
\begin{center}
\begin{tabular}{|c |c c c |c c  c|}
\hline
%\multicolumn{1}{c|}{}& \bf A\_2S\_bef\_200K &  \bf ABC\_2S\_bef\_600K \\
\bf Data & \multicolumn{3}{c|}{\bf 200K-D} \\
\bf \# examples & \multicolumn{3}{c|}{200K} \\
 \bf Percentiles & $P_5$ & $P_{50}$ & $P_{95}$ \\
\hline \hline
\bf RegMLP & 0.6 & 1.3 & 1.9 \\ % exp1 xx, exp13 xx(svp pr√©server ces commentaires) 
\hline
\bf HeurV & 0.3 & 1.0 & 1.6  \\ 
\bf HeurS & 1.9 & 3.9 & 4.1 \\ %
% \bf RegMLP & \num{7.4e-4} & \num{1.5e-3} & \num{2.3e-3} & \num{5.6e-4} & \num{1.3e-3} & \num{1.9e-3} \\ % exp1 xx, exp13 xx(svp préserver ces commentaires) 
% \hline
% \bf HeurV & \num{3.0e-4} & \num{9.8e-4} & \num{1.6e-3} & \num{3.0e-4} & \num{9.9e-4} & \num{1.6e-3}  \\ 
% \bf HeurS & \num{1.9e-3} & \num{2.9e-3} & \num{4.0e-3} & \num{1.9e-3} & \num{3.9e-3} & \num{4.1e-3} \\ %
\hline
\end{tabular}%timesT
\caption{Prediction time in milliseconds per instance with extraneous data}
\label{table:timesTestingOutside}
\end{center}
%\end{scriptsize}
\end{table}

\section{Conclusion and Future Work} \label{sec:conclusion}

This paper has proposed a supervised ML approach for predicting expected TDOSs under imperfect information in short computing time. The problem is of relevance to various applications where tactical and operational planning problems are interdependent. We considered an application related to railway demand and capacity management at the tactical level (accept / reject booking requests) whose solution depends on a kind of packing problem at the operational level. A similar problem occurs in other freight transportation settings, for example, airline cargo and less-than-truckload.

We formulated the problem as an optimal prediction stochastic programming problem whose solutions we predicted with machine learning. Key in the proposed methodology is the generation of labeled training data for supervised learning. We proposed to sample operational problem instances (perfect information) by controlled probabilistic sampling. The generated operational problem instances were solved independently and offline using an existing solver. We handled uncertainty with appropriate aggregation methods. Otherwise, our methodology relies on existing ML models and algorithms. This is a substantial advantage since we can benefit from the recent advances in ML, and in our case, deep learning.

We illustrated the methodology with a train LPP, where some features of the inputs (problem instances) -- in this case container weights -- are unavailable at the time of prediction. We explained why the methodology cannot be viewed as supplying heuristic solutions to the LPP.
The results showed that a regression feedforward neural network presented the best performance overall. Remarkably, the solutions could be predicted with a high accuracy in comparison with an error bound, and in very short computing time (in the order of a millisecond or less). In fact, the time required to predict the solution descriptions under imperfect information using ML is much shorter than the time required to solve a single deterministic instance with an ILP solver. The results also showed that the regression feedforward neural network model that was trained and validated on simpler instances could generalize surprisingly well to harder instances without specific training and validation. However, quite expectedly, the variations over the hyperparameter sets considered during the validation step were large when the nature of the data was very dissimilar.

We considered an input and output structure of small and fixed size. A direction for future research is to predict more detailed solutions where the input and output structures would be of large and variable size and would possibly feature additional constraints. The trade-off between the level of detail and uncertainty of the input is a question by itself. In this context, an approach related to pointer networks \citep{VinyEtAl15} might be a promising avenue. Since data generation is the most expensive part of the methodology, future research should investigate learning algorithms whose trade-off between the cost of generating data and the predictive performance can be controlled.

Finally, we envision that the same methodology could be successfully applied to other types of two-stage stochastic programming problems like, for example, the two-stage Vehicle Routing Problem (VRP) with stochastic demands as defined by \cite{Gendreau14}. 
In this context, the routes are specified in the first stage and the intermediate returns to the depot are determined online in the second stage. This framework gives rise to two formulations, respectively based on network flows and on set partitioning. Given a particular vehicle route, the operational solution consists of all two-way trips made to the depot from the customer locations when the capacity of the vehicle is exceeded. In the network flow and the set partitioning formulations, given a particular vehicle route, the expected tactical description of operational solution could be the probability of a return to depot at a specific location along this route \citep[][Eq. (8.9)]{Gendreau14} and the cumulative probability distribution of the cumulative demand at a location along this route \citep[][p. 219]{Gendreau14}, respectively. Predicting either of the latter with ML would provide advance information on the operational solution that would be valuable from a tactical standpoint and possibly accelerate the computation of the expected recourse function and the online computation of the solution to the overall VRP problem.

\section*{Acknowledgments}

This research was funded by the Canadian National Railway Company (CN) Chair in Optimization of Railway Operations at Universit\'e de Montr\'eal and a Collaborative Research and Development Grant from the Natural Sciences and Engineering Research Council of Canada (CRD-477938-14). Computations were made on the supercomputers Briar\'ee and Guillimin, managed by Calcul Qu\'ebec and Compute Canada. The operation of these supercomputers is funded by the Canada Foundation for Innovation (CFI), the Minist\`ere de l'\'Economie, de la Science et de l'Innovation du Qu\'ebec (MESI) and the Fonds de recherche du Qu\'ebec - Nature et technologies (FRQ-NT). The research is also partially funded by the ``IVADO Fundamental Research Project Grants'' under project entitled ``Machine Learning for (Discrete) Optimization''. We are grateful for important insights obtained through discussions with Jean-Fran\c{c}ois Cordeau, Matteo Fischetti and Michael Hewitt. We are indebted to three anonymous referees for extremely careful readings and for challenging us to clarify the essence of the contribution.

\bibliographystyle{plainnat_custom}

\bibliography{References}

\end{document}